\pdfoutput=1

\documentclass[11pt]{article}

\usepackage{EMNLP2023}

\usepackage{times}
\usepackage{latexsym}
\usepackage{booktabs}
\usepackage{graphicx}
\usepackage[T1]{fontenc}

\usepackage[utf8]{inputenc}

\usepackage{microtype}

\usepackage{inconsolata}

\usepackage{mathtools}
\usepackage{mplcolors}

\title{``Honey, Tell Me What's Wrong'', Global Explanation of Textual Discriminative Models through Cooperative Generation}

\author{Antoine Chaffin$^{*}$ \\
  IRISA, Rennes, France \\
  IMATAG, Rennes, France \\
  \texttt{antoine.chaffin@irisa.fr} \\\And
  Julien Delaunay$^{*}$\\
  IRISA, Rennes, France \\
  Inria, Rennes, France \\
  \texttt{julien.delaunay@inria.fr} \\}

\begin{document}
\maketitle
\begin{abstract}
    The ubiquity of complex machine learning has raised the importance of model-agnostic explanation algorithms. These methods create artificial instances by slightly perturbing real instances, capturing shifts in model decisions. However, such methods rely on initial data and only provide explanations of the decision for these. To tackle these problems, we propose Therapy, the first global and model-agnostic explanation method adapted to text which requires no input dataset. Therapy generates texts following the distribution learned by a classifier through cooperative generation. Because it does not rely on initial samples, it allows to generate explanations even when data is absent (e.g., for confidentiality reasons). Moreover, conversely to existing methods that combine multiple local explanations into a global one, Therapy offers a global overview of the model behavior on the input space. Our experiments show that although using no input data to generate samples, Therapy provides insightful information about features used by the classifier that is competitive with the ones from methods relying on input samples and outperforms them when input samples are not specific to the studied model.
\end{abstract}
\def\thefootnote{*}\footnotetext{Equal contribution.}
\section{Introduction}
The emergence of machine learning models has led to their adoption in domains spanning from mere recommendations to critical areas such as healthcare~\cite{ia_medicine,breast_cancer_prediction} and law~\cite{law_overview}. These already complex models keep becoming larger, emphasizing their black-box denomination. This lack of transparency however slows their adoption in various areas since we witness a notable rise of deployed models suffering from bias. For example, some chatbots biased toward religious~\cite{gpt3-muslim} and gender~\cite{gpt3-gender} minorities have been released and explaining their inner mechanisms is still an ongoing problem. 

Among the methods proposed to tackle these problems, model-agnostic approaches are favored since applicable to any machine learning model. Among these, local explanations have obtained strong success by maintaining a good trade-off between accuracy and transparency. These explanations are generated in the proximity of a target instance by tampering this input to create neighbors and study how the model reacts to these changes. This allows them to highlight which features are important for the model and to provide explanations on the decision for this input (e.g., the most important words for each class). According to a recent study~\cite{trends_in_xai}, LIME~\cite{DBLP:conf/kdd/Ribeiro0G16}, while being the first model-agnostic local explanation method is still the most widely used. However, local explanations have three main flaws when trying to explain a model. First, it obviously requires to have inputs to explain, which might not be possible due to confidentiality or privacy reasons~\cite{amin-nejad-etal-2020-exploring}. Second, selecting inputs that are representative of the model or the downstream data distribution is difficult. Finally, it will explain the decision \textbf{for this input} and for this input only. This only provides very local information on the model behavior, which represents only a very small piece of the input domain of the model. Therefore, LIME and other local explanation methods have proposed to aggregate the information from multiple samples to provide global explanations. However, these explanations are strongly tied to the input samples and only provide cues about the samples' neighborhood. These methods thus require samples that cover as much of the space as possible.

To relax this sample dependency and generate global explanations of the model, we propose \textbf{Therapy}, a method that leverages cooperative generation~\cite{DBLP:conf/acl/ChoiBGHBF18, DBLP:conf/icml/ScialomDLPS20, DBLP:conf/iclr/DengBOSR20,DBLP:conf/naacl/ChaffinCK22} to generate texts following the distribution of a classifier. The distribution of the resulting samples can then be used to study which features are important for the model, providing global information on its behavior.

In this paper, we first introduce the related work in Section~\ref{sec:related_work} and cooperative text generation in Section~\ref{sec:text_generation}. We then present Therapy in Section~\ref{sec:method} and the experiments conducted to compare its performance to standard explanation methods in Section~\ref{sec:experiments}.

\section{Related work}
\label{sec:related_work}
Generating explanations for textual data is challenging since it requires considering both the text semantics and task domains. Moreover, it is frequent that models are already deployed and further evaluations are required (e.g., fairness, bias detection) but the training data is not accessible. This may be caused by data privacy, security, or simply because the dataset is too large to be analyzed. Thus, to fulfil this objective, researchers have focused on post-hoc explanations~\cite{trends_in_xai}. Following the categorization by Bodria et al.~\cite{xai_survey_pisa}, we distinguish between example-based and feature-attribution explanations.

\subsection{Example-Based Explanations}
Taking roots from social science~\cite{miller}, the example-based explanations indicate either the minimum change required to modify the prediction --counterfactual-- or illustrate class by showing representative instances --prototypes--. Counterfactual methods answer "what if" questions and have gained interest since being close to human reasoning, perturbing document until the model prediction differs~\cite{wachter}. Conversely, prototype methods select or generate representative instances for the target class. Among the example-based methods, some leverage on control codes to perturb the input text while others generate realistic sentences based on perturbation in a latent space. Polyjuice~\cite{polyjuice} and GYC~\cite{gyc} belong to the former and propose control codes varying from changing the sentiment and tense of the sentence to adding or replacing words. On the other hand, xSPELLS~\cite{xspells} and CounterfactualGAN~\cite{cfgan} are methods that train respectively a Variational Autoencoder and a Generative Adversarial Network to convert input text to a latent space and return realistic sentences from this latent space. These methods hence convert the input document into a latent space and slightly perturb it until the closest counterfactual is found.

\subsection{Feature-Attribution Explanations}
Feature-attribution methods assign weights to input words, indicating the positive or negative impact on the final prediction. Methods such as SHAP~\cite{shap}, LIME~\cite{DBLP:conf/kdd/Ribeiro0G16}, and their variants~\cite{slime,DLIME,OptiLIME,iLIME,QLIME} are the most commonly used~\cite{trends_in_xai}. They are local since they perturb an input instance by slightly modifying it and studying the complex model in a given locality. For textual data, LIME randomly masks the words of the input document and trains a linear model on the collection of perturbed documents to predict the decisions of the complex model. The most important coefficients of the linear model associated with the input words are then returned as the explanation. While most explainability surveys~\cite{survey_XAI,xai_survey_pisa} differentiated between local and global explanations, LIME also introduced LIME-SP (for submodular pick), a global method that generates $n$ local explanations for a set of individual instances. These $n$ instances are selected to cover as much of the input domain as possible and avoid redundancy.

\section{Text generation}
\label{sec:text_generation}
\subsection{Cooperative Generation}
Language Models (LM) such as the GPT family~\cite{Radford2018ImprovingLU, Radford2019LanguageMA, DBLP:conf/nips/BrownMRSKDNSSAA20} learn the probability distribution of sequences of symbols $x_1, x_2, \cdots, x_T$ (most often \textit{tokens}) taken from a vocabulary $\mathcal{V}$, with variable lengths $T$. The probability of one sample $x$ (also called \textit{likelihood}) is defined as the joint probabilities over each of its tokens, which can be factorized using the chain rule: $p(x_{1:T})=\prod_{t=1}^{T} p(x_{t} \mid x_{1:t-1})$. The LM is trained to output a probability distribution over the dictionary for the next token given the input ones i.e.  $p(x_t \mid x_{1:t-1})$ at a given time step $t$. This results in an auto-regressive LM that can generate sequences by iteratively using those distributions to emit a token $x_t$, and append it to the context $x_{1:t-1}$ for the next iteration. The generation process --or \textit{decoding}-- is often started using a small initial sequence: the \textit{prompt}. 
Large LMs learn an excellent approximation of the true distribution of their training data, so generating samples that maximize the model likelihood $p(x)$ allows to generate plausible texts. However, this approach offers very little control over the text being generated besides the initial prompt.

Cooperative generation approaches~\cite{DBLP:conf/acl/ChoiBGHBF18, DBLP:conf/icml/ScialomDLPS20, DBLP:conf/iclr/DengBOSR20}, where discriminative models are used to guide the LM during the generation, offer more control. They use the information from the external model to guide the LM to generate texts that have a property it recognizes. In situations where the model is a classifier which learns to output the probability $D(c \mid x)$ of a sequence $x$ to belong to a class $c$, the goal is to generate text that maximizes the probability of belonging to the target class. Evaluating $D(c \mid x)$ for every sequence possible is intractable due to the size of the space ($|\mathcal{V}|^n$ for a sequence of length $n$). Thus, these methods leverage the distribution of the LM to restrict the exploration to plausible sequences. This results in a sequence that is both well written and belongs to the target class since the produced sequence maximizes $p(x)*D(c \mid x) \propto p(x \mid c)$.

\subsection{Monte Carlo Tree Seach Guided Decoding}
\label{ssec:mcts}
Among cooperative approaches, the ones that leverage the 
Monte Carlo Tree Search (MCTS) to guide the decoding of the LM exhibited very strong results~\cite{selfGAN, DBLP:conf/naacl/ChaffinCK22, DBLP:conf/emnlp/LeblondASPLASV21, DBLP:conf/icml/LamprierSCCKSP22}. 
MCTS is an iterative algorithm that seeks solutions in a tree space too large to be exhaustively searched. It is applicable to text generation because the search space created during decoding corresponds to a tree: the prompt is the root and the children of a node are its parents' sequence with one additional token. MCTS loop is composed of four steps: selection, expansion, simulation and back-propagation.

\begin{enumerate}
    \item \textbf{Selection} An exploration from the root of the tree to an unexplored leaf. The path to the leaf is defined by selecting, at each node, the children that maximize the Polynomial Upper Confidence Trees (PUCT)~\cite{DBLP:journals/amai/Rosin11, DBLP:journals/nature/SilverSSAHGHBLB17}, which is, for a node $i$: 
    \begin{equation*}
        \label{eqn:puct}
        PUCT(i) = \frac{s_{i}}{n_{i}} + c_{puct} \; p(x_i \mid x_{1:t-1})\frac{\sqrt{N_{i}}}{1+n_i}
    \end{equation*}
    with $n_{i}$ the number of simulations played after the node $i$, $s_i$ its aggregated score, $N_{i}$ the number of simulations played after its parent, and $c_{puct}$ a constant defining the compromise between exploitation (focusing on nodes with already good scores) and exploration (exploring promising nodes).
    
    \item \textbf{Expansion.} The creation of the selected node children if it is not terminal (i.e., corresponding to the end-of-sequence token).
    
    \item \textbf{Simulation (roll-out).} The sampling of additional tokens (using the LM distribution) until a terminal node.
    
    \item \textbf{Back-propagation.} The evaluation of the sequence $x$ associated with the terminal node and aggregation of its score to each parent until root. In order to guide the generation towards texts that belong to a given class according to a classifier, the score of the sequence $x$ associated with a given leaf can be defined as $D(c \mid x)$ given by the classifier. Different aggregation strategies can be used, such as computing the average of the actual score of the node and the terminal node one as in ~\cite{DBLP:conf/naacl/ChaffinCK22} or taking the maximum of the two as in ~\cite{DBLP:conf/nips/ScialomDSLP21, DBLP:conf/icml/LamprierSCCKSP22}.
\end{enumerate}

This loop is repeated a given number of times (defining the compute budget) and the tree produced is then used to select the token to add for the current decoding step. It can be selected as the most played node among the root’s children nodes, or the one with the highest aggregated score. Since we are interested in generating sequences that are as stereotypical of classes of the discriminative model as possible, we choose the node with the highest score. The selected node then becomes the new root and the process is repeated until the final sequence is produced.

\begin{figure*}
\centering
\includegraphics[width=0.8\textwidth]{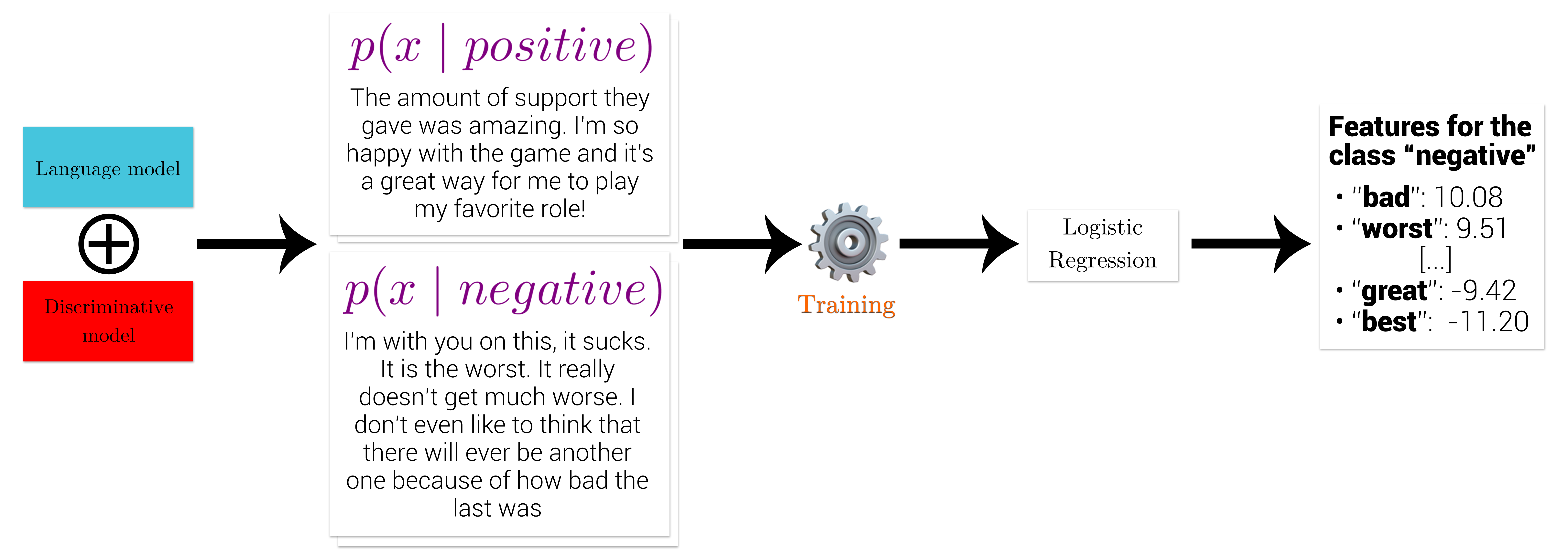} 
    \caption{Illustration of the Therapy method. Texts from different classes are cooperatively generated using the guidance of the studied model. A logistic regression is then trained to predict the label of the generated texts. The weights of the model associated with each word are then returned as importance weights.}
    \label{fig:therapy}
\end{figure*}

Contrary to traditional left-to-right decoding strategies that can miss sequences that gets better after some steps or be trapped in sub-optimal sequences, MCTS breaks the myopic decoding by defining the score of a token based on possible continuations of the sequence.
In addition to being plug-and-play, i.e, any type of (auto-regressive) language model can be guided during decoding by any type of classifier using MCTS, this approach exhibited state-of-the-art results in the task of constraint generation, that is, generating texts that maximize $D(c \mid x)$ while maintaining a high quality of writing. We thus experiment with MCTS decoding for Therapy, but the proposed method is compatible with any cooperative generation approach.

\section{Method}
\label{sec:method}
In this paper, we introduce \textbf{Therapy}, a global and model-agnostic explanation method that does not require input data. In place of these input data, Therapy employs an LM guided by the model to explain. This cooperation generates texts that are representative of the classes learned by the studied discriminative model. To do so, Therapy extracts the most important words for the classifier by employing it to steer an LM through cooperative generation. Texts generated using cooperative generation follow the distribution $p(x) * D(c \mid x)$. Their distribution can thus be used to study the classifier $D$: words with high frequencies are likely to be important for the classifier. A logistic regression is then learned on tf-idf representations of generated samples and the weights associated with each term are returned as the explanation. An illustration of the method is proposed in Figure~\ref{fig:therapy}. Because $p(x)$ is the same for every class, by using tf-idf on the whole corpus (i.e., samples from every class), words that are frequent because of $p(x)$ or in multiple classes will be filtered out. Hence, the logistic regression model learned on the tf-idf score of each feature allows Therapy to study their relative importance and to extract the most important ones for each class. The method thus offers the level of explainability of n-grams based on logistic regression models to any classifier. Indeed, since any type of (auto-regressive) LM can be guided during decoding by any classifier using MCTS, the proposed approach is totally model-agnostic. 

We call this approach Therapy because its functioning is similar to that of a therapist. This therapist (the LM) queries its patient (the classifier) to understand its behavior and eventually discover pathologic behaviors (some biases). 

In essence, the method is similar to using LIME jointly with a masked LM to generate neighbors when the number of replaced tokens grows a lot but with two benefits.
First, the method does not rely on input examples but creates samples out of nothing using the LM. This is useful for cases where the data cannot be shared because it contains confidential information~\cite{amin-nejad-etal-2020-exploring}. Moreover, rather than exploring the neighborhood of these examples (and so conditioning the explanations on these examples' context), the domain of the exploration is defined by the domain of the LM, which is significantly broader. Besides, either a general LM can be used to study the model behavior on generic data or an LM specific to the downstream domain to make sure it works well on this specific type of data. 

Second, the method does not generate \textbf{before} classifying the text but employs the classifier \textbf{during} the generation. Hence, instead of "randomly" generating texts and hoping for important features to appear, we explicitly query the model for stereotypic features by maximizing $D(c \mid x)$. This makes the method more efficient and reduces the probability of generating rare features that are not important for the model while reducing the odds of generating "in the middle" texts containing features from various classes that are misleading.
Besides, our method directly relies on the distribution learned by the studied model to guide the generation, unlike methods like Polyjuice and GYC, which, in addition to requiring input data, count on a distribution learned by the LM to bias the generation towards the desired property (using control codes).

Finally, Therapy is distinctive from methods analyzing the frequency of input terms in the training data such as sensitivity analysis since it does not require access to (training) data and directly exploits the distribution effectively learned by the model, whereas nothing guarantees that a model is actually using the terms extracted from training data to make a prediction. Furthermore, our method differs from existing example-based and feature attribution methods since to the best of our knowledge, there exists no global and model-agnostic explanation methods that do not require any input data.

\begin{figure*}[ht]
    \centering 
    \includegraphics[width=0.8\textwidth]{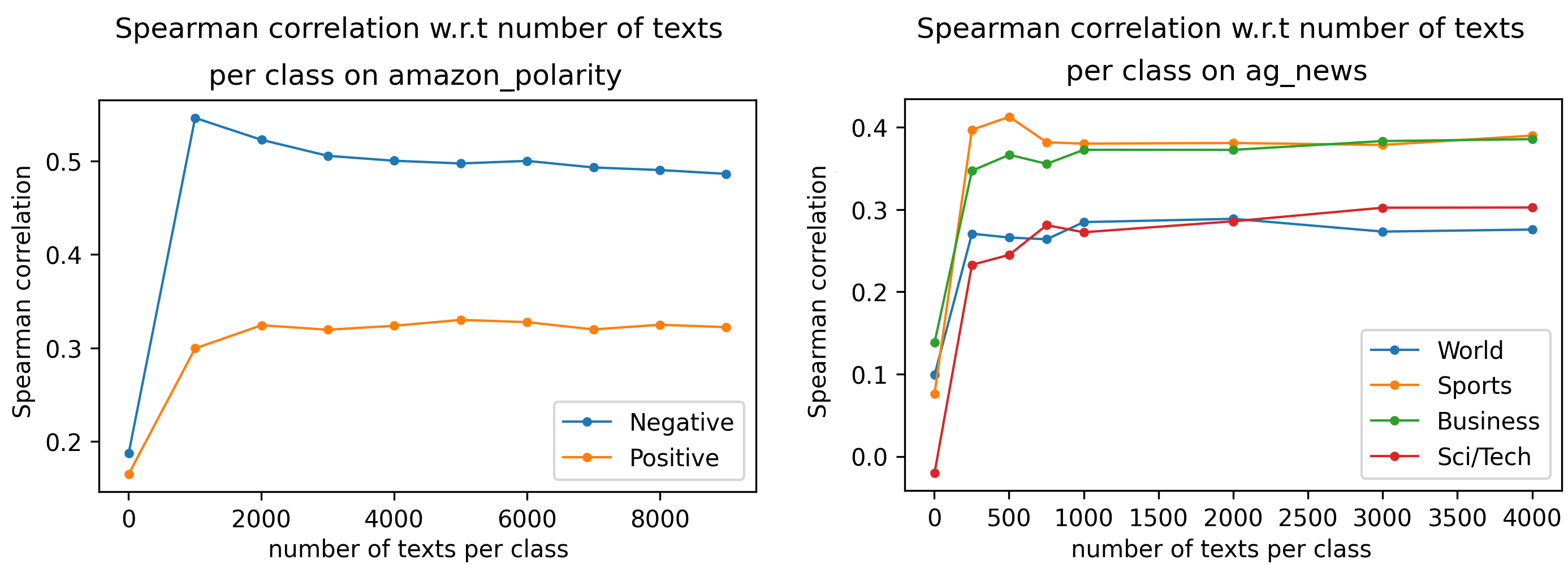} 
    \caption{Spearman correlation w.r.t number of generated text per class for amazon\_polarity and ag\_news.}
    \label{fig:spearman_wrt_generated}
\end{figure*}

\section{Experiments}
\label{sec:experiments}
In this section, we first give technical details on the experiments conducted to evaluate Therapy (Section~\ref{subsec:configuration}). We then evaluate Therapy through three experiments. The first one (Section~\ref{subsec:spearman_correlation}), measures the Spearman correlation of the explanations and the weights of a glass box and studies the influence of the number of generated texts on the quality of the explanation returned by the linear model. We then compare the capacity of the method to correctly identify the most important words of the glass box to the one of LIME and SHAP using precision/recall curves in Section~\ref{subsec:precision_recall}. Finally, we test whether the terms returned by the different approaches are sufficient to modify the prediction of the classifier in Section~\ref{subsec:insertion_suppression}. The code of Therapy and our experiments will be made available upon acceptance.

\subsection{Experimental setup}
\label{subsec:configuration}
\paragraph{Glass-box explanation}
Since there are no ground truth explanations available to be used as a goal for evaluated methods, we use a glass-box model, that is, a model explainable by design but used as a black box (i.e., without being able to use its inner workings to generate explanations). Following prior work~\cite{DBLP:journals/ai/Guidotti21}, we train a logistic regression using \href{https://scikit-learn.org/stable/modules/generated/sklearn.linear_model.LogisticRegression.html}{sklearn}~\cite{scikit-learn} and use its weights as tokens importance scores. 

\paragraph{Therapy implementation}
To evaluate the proposed method, we use the available implementation of \href{https://github.com/NohTow/PPL-MCTS}{PPL-MCTS}~\cite{DBLP:conf/naacl/ChaffinCK22} and simply plug the glass-box by defining the function that takes a sequence and returns its score. The choice of the LM to guide defines the domain on which we want to explain the behavior of the model. Thus, it is best to choose a language model that is as close as the domain on which the discriminator will be used. However, to show that the proposed approach works well, even on a general domain, we use \href{https://huggingface.co/facebook/opt-125m}{OPT-125m}~\cite{DBLP:journals/corr/abs-2205-01068}. A logistic regression is then learned on generated texts and its scores are used as token importance.

\paragraph{Datasets}
Experiments are conducted on two different classification datasets from~\cite{DBLP:conf/nips/ZhangZL15}. The first one, \href{https://huggingface.co/datasets/amazon_polarity}{amazon\_polarity} is a binary classification dataset of Amazon reviews labelled as positive or negative. The reviews are rather small and have highly caricatural lexical fields. The second one, \href{https://huggingface.co/datasets/ag_news}{ag\_news}, is a thematic classification dataset with 4 classes: (\texttt{world}, \texttt{sport}, \texttt{business} and \texttt{sci/tech}). Texts in this dataset are longer and more diverse but include distinctive indicators because they are extracted from online news articles. Samples generated by Therapy along with top words returned by the method for each class of both datasets are given in Appendix~\ref{sec:qualitative_results}.


\begin{table*}
    \centering
    \resizebox{\textwidth}{!}{
    \begin{tabular}{lcccccc}
         Dataset &    \multicolumn{2}{c}{\textsc{amazon\_polarity}} &    \multicolumn{4}{c}{\textsc{ag\_news}}   \\ 
         \cmidrule(r){2-3} \cmidrule(r){4-7}
         Class & Positive & Negative  & World & Sports & Business & Sci/Tech  \\
        \midrule
         Baseline &   0.49 (6.24e-08) & 0.31 (9.25e-05)    &   0.25 (1.67e-06)    & 0.32 (6.58e-09) &  0.35 (1.88e-11)   & 0.12 (2.33e-02) \\
        \midrule
        Therapy - most played &   0.52 (5.79e-09) & 0.32 (7.83e-05)    &   0.22 (1.57e-05)    & 0.27 (7.66e-07) &  0.32 (2.04e-09)   & 0.22 (1.93e-05) \\
        \midrule
        Therapy - highest score &   0.49 (3.3e-08) & 0.31 (1.0e-04)    &   0.27 (1.6e-07)    & 0.37 (4.0e-12) &  0.38 (5.6e-13)   & 0.3 (8.9e-09) \\
    
    \end{tabular}
    }
    \caption{Spearman correlation (p-value) between the top words of a logistic regression glass-box and explanation methods learning a logistic regression over generated texts. Baseline uses unconstrained samples while Therapy generates samples using the MCTS, either selecting the most played or highest scored node. Results are shown per class and dataset.}
    \label{tab:classifier_guidance}
\end{table*}

\paragraph{Compared methods}
In our experiments, we compare the results of Therapy to the two most widely used post-hoc methods: \href{https://github.com/marcotcr/lime}{LIME}~\cite{DBLP:conf/kdd/Ribeiro0G16} and \href{https://shap.readthedocs.io/en/latest/}{SHAP}~\cite{shap}. We employed publicly available implementations of these traditional methods instead of their extensions mentioned in Section~\ref{sec:related_work}. This decision was made because, to the best of our knowledge, these extensions either do not prioritize the generation of global explanations or do not enhance the textual versions of these methods. The main difference between LIME and SHAP is that the former generates samples by modifying input data and then learns a linear regression model whereas the latter benefits from game theory to compute the weight of each term. We use the global version of these methods on 500 texts of the datasets test set. For SHAP, we keep the 10 000 most important words for each dataset whereas, for LIME, we computed 500 local explanations with the 35 most important words and merged every term-weights pair into dictionaries of length 4592 for amazon\_polarity and 5770 for ag\_news.
Finally, to highlight the benefits of cooperative generation in Therapy, we also report the results obtained by a simple baseline. Rather than using cooperatively generated texts to train the logistic regression, the baseline generates texts without constraining the LM and uses the glass-box \textbf{after} the generation is done to get the target labels.

\begin{table*}[]
    \centering
    \resizebox{\textwidth}{!}{
    \begin{tabular}{lcccccc}
         Dataset &    \multicolumn{2}{c}{\textsc{amazon\_polarity}} &    \multicolumn{4}{c}{\textsc{ag\_news}}   \\ 
         \cmidrule(r){2-3} \cmidrule(r){4-7}
         Class & Positive & Negative  & World & Sports & Business & Sci/Tech  \\
        \midrule
         Baseline &   0.49 (6.24e-08) & 0.31 (9.25e-05)    &   0.25 (1.67e-06)    & 0.32 (6.58e-09) &  0.35 (1.88e-11)   & 0.12 (2.33e-02) \\
        \midrule
        LIME & 0.64 (5.0e-7) & 0.44 (1.5e-3)  &  0.09 (0.53) & 0.16 (0.27) & 0.20 (0.16) & 0.19 (0.19) \\
        \midrule
        LIME-other & 0.21 (0.14) & 0.18 (0.21) & -0.03 (0.85) & 0.23 (0.12) & 0.09 (0.52) & 0.29 (0.04) \\
        \midrule
        SHAP & 0.71 (7.6e-9)  & 0.76 (1.6e-10)  & 0.47 (6.2e-4)  &  0.62 (1.7e-06) &  0.53 (8.0e-5)  &   0.61 (2.4e-6)\\
        \midrule
        SHAP-other & 0.02 (0.87) & 0.26 (0.06) & -0.05 (0.71) & 0.04 (0.77) & 0.15 (0.31) & 0.12 (0.41) \\
        \midrule
        Therapy &   0.49 (3.3e-08) & 0.31 (1.0e-04)    &   0.27 (1.6e-07)    & 0.37 (4.0e-12) &  0.38 (5.6e-13)   & 0.3 (8.9e-09) \\
       
    \end{tabular}
    }
    \caption{Spearman correlation (p-value) between the top words of a logistic regression glass-box and the four explanation methods. `other' indicates that the explanations are generated using the other dataset. Results are shown per class and dataset.}
    \label{tab:spearman}
\end{table*}

\subsection{Spearman correlation}
\label{subsec:spearman_correlation}
A good explanation of the glass box is a list of features that contains both its important features (i.e., has good coverage) and links them to a similar relative weight. Hence, we compute the Spearman correlation between the top words of the glass box (having a weight $>1$) and their scores attributed by the explainer. 
We selected Spearman correlation over Pearson because the score returned by LIME and SHAP can be very different from logistic regression weights and so rank correlation results in a fairer comparison.

\subsubsection{Influence of the number of generated texts}
\label{subsec:number_of_text}
One critical parameter of the proposed method is the number of texts to generate since more tokens allow a larger coverage but require more computation. We report the Spearman correlation against the number of generated texts per class in Figure~\ref{fig:spearman_wrt_generated}. We observe that the correlation quickly rises until plateauing, meaning that only a small amount of text offers a great overview of the model behavior and that the method does not require a lot of computing to perform. We thus fixed the number of generated texts for Therapy to 3000 for each class for the rest of our experiments.

\subsubsection{Importance of the classifier guidance}
Cooperative generation allows Therapy to guide the LM during the decoding process and to move away from its distribution toward that of the model studied. To study the importance of this guidance, we report, in addition to the baseline, the results obtained when selecting the most played token during MCTS generation. As mentioned in Section~\ref{ssec:mcts}, the token added to the current context can be selected as the most played node or the one obtaining the highest score. Selecting the highest-scored node generates texts that are the most stereotypical of the studied model, while the most played node is closer to the LM a priori. Results reported in Table~\ref{tab:classifier_guidance} show that both the baseline and using the most played node exhibit competitive results on amazon\_polarity but struggle more on ag\_news. This can be explained by the fact that the LM tends to not generate positive and negative terms at the same time, so the classes are clearly defined even in unconstrained samples. On ag\_news, however, there is more overlap between classes, and so using cooperative generation helps to generate texts that are more distinctive of a given class. These results both highlight the contribution of the cooperative generation and motivate the token selection method.

\subsubsection{Comparison with other methods}
The Spearman correlations of all the evaluated approaches can be found in Table~\ref{tab:spearman}.
Results yielded by Therapy are better than those of LIME on ag\_news but worse on amazon\_polarity whereas SHAP yields better results than both methods on both datasets. Counterintuitively, these are positive results for Therapy because other methods have access to the test set of the studied dataset, ensuring that the target features are found in the input examples. To test the performance when this assumption no longer holds, we resort to two variants of LIME and SHAP, denoted by \textit{-other}. The key distinction between these methods lies in the dataset employed as input data. We use amazon\_polarity texts as input to find features in ag\_news and vice-versa. The findings from these experiments reveal that existing methods fail to find important features, leading to a significant drop in correlations, substantially lower than those of Therapy. 

\subsection{Precision Recall}
\label{subsec:precision_recall}
\begin{figure*}
\includegraphics[width=\textwidth]{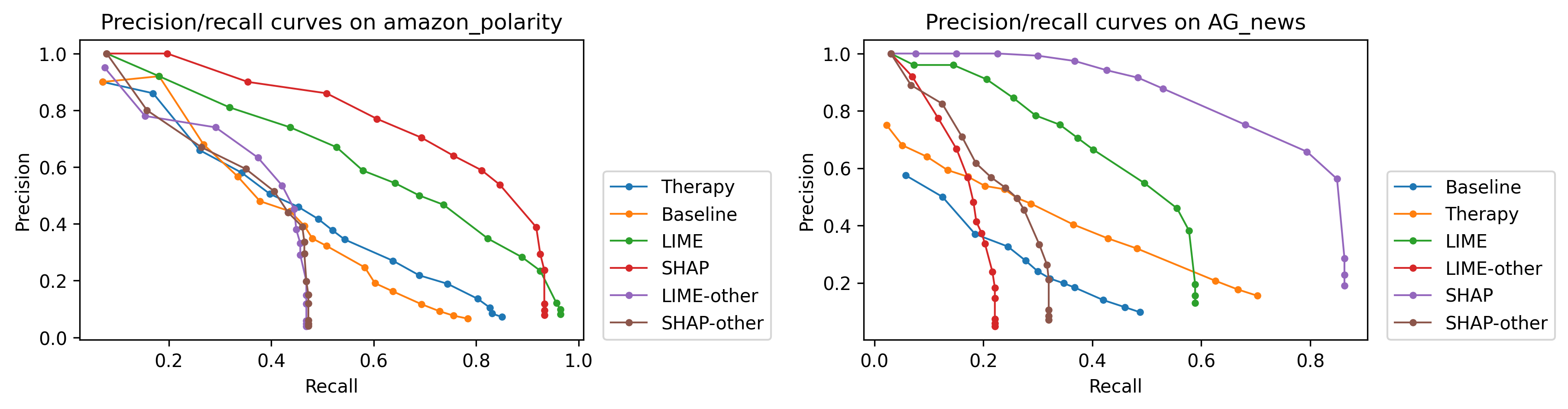} 
    \caption{Precision/recall curves of the glass-box top words for the different explanation methods.}
    \label{fig:recallprec}
\end{figure*}
Besides assigning correct scores to important features of the model, we also want to make sure that Therapy gives an informative output in practice. That is, making sure that most features returned by the explainer (i.e., its highest-scored features) are indeed important features of the original model and that most of its important features are found. Thus, we report precision/recall curves averaged over every class in Figure~\ref{fig:recallprec}. Precision is obtained by computing, for different numbers of words returned, the proportion that is in the most important features of the original model. Conversely, recall is the proportion of the original model's top words retrieved. The number of words returned ranges from 10 to 1500.

\begin{figure*}[ht]
    \centering
    \begin{minipage}{.49\textwidth}
      \centering
            \includegraphics[width=\columnwidth]{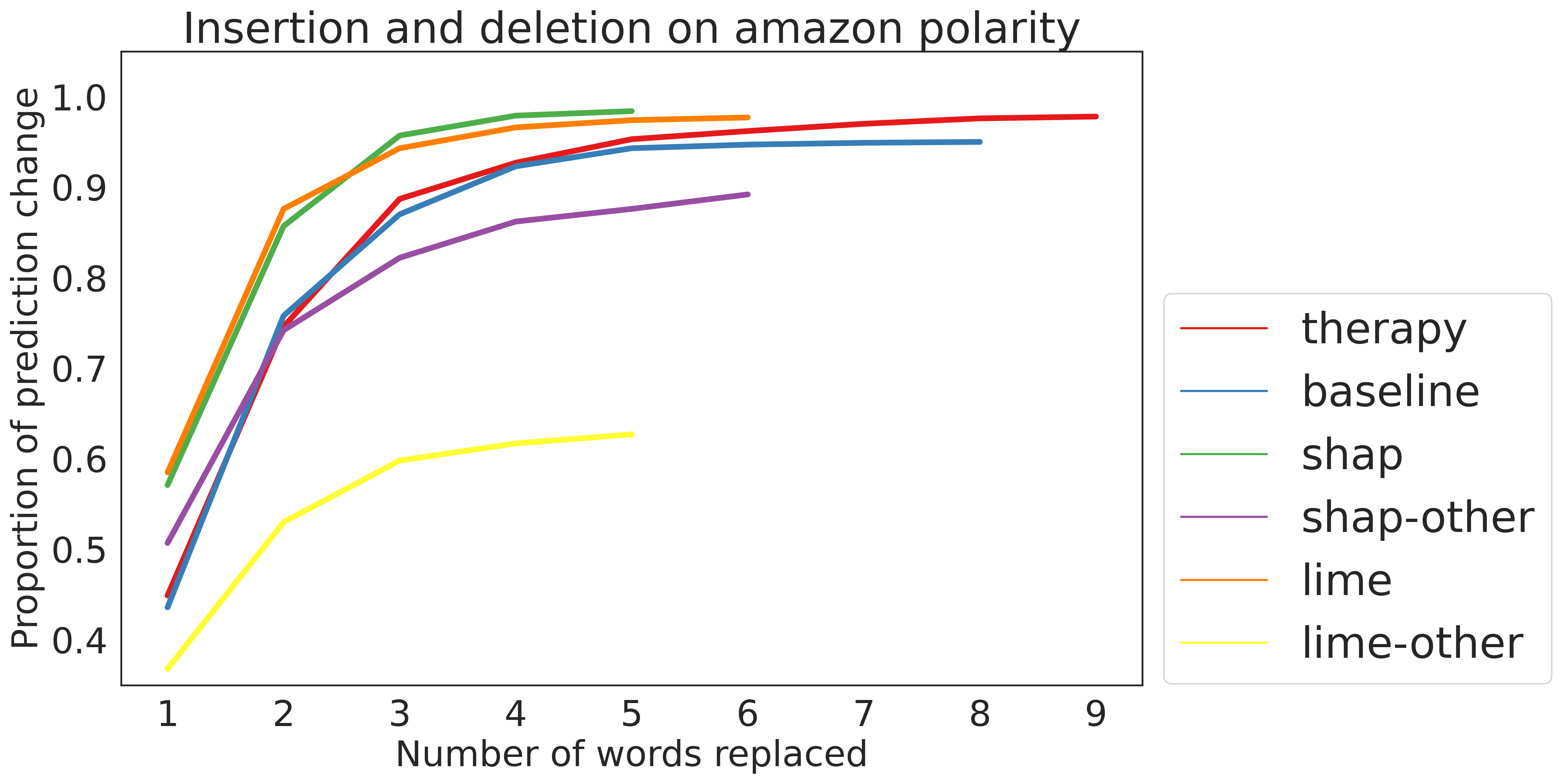}
      \label{fig:test1}
    \end{minipage}%
    \hspace{0.5em}    
    \begin{minipage}{.49\textwidth}
      \centering
        \includegraphics[width=\columnwidth]{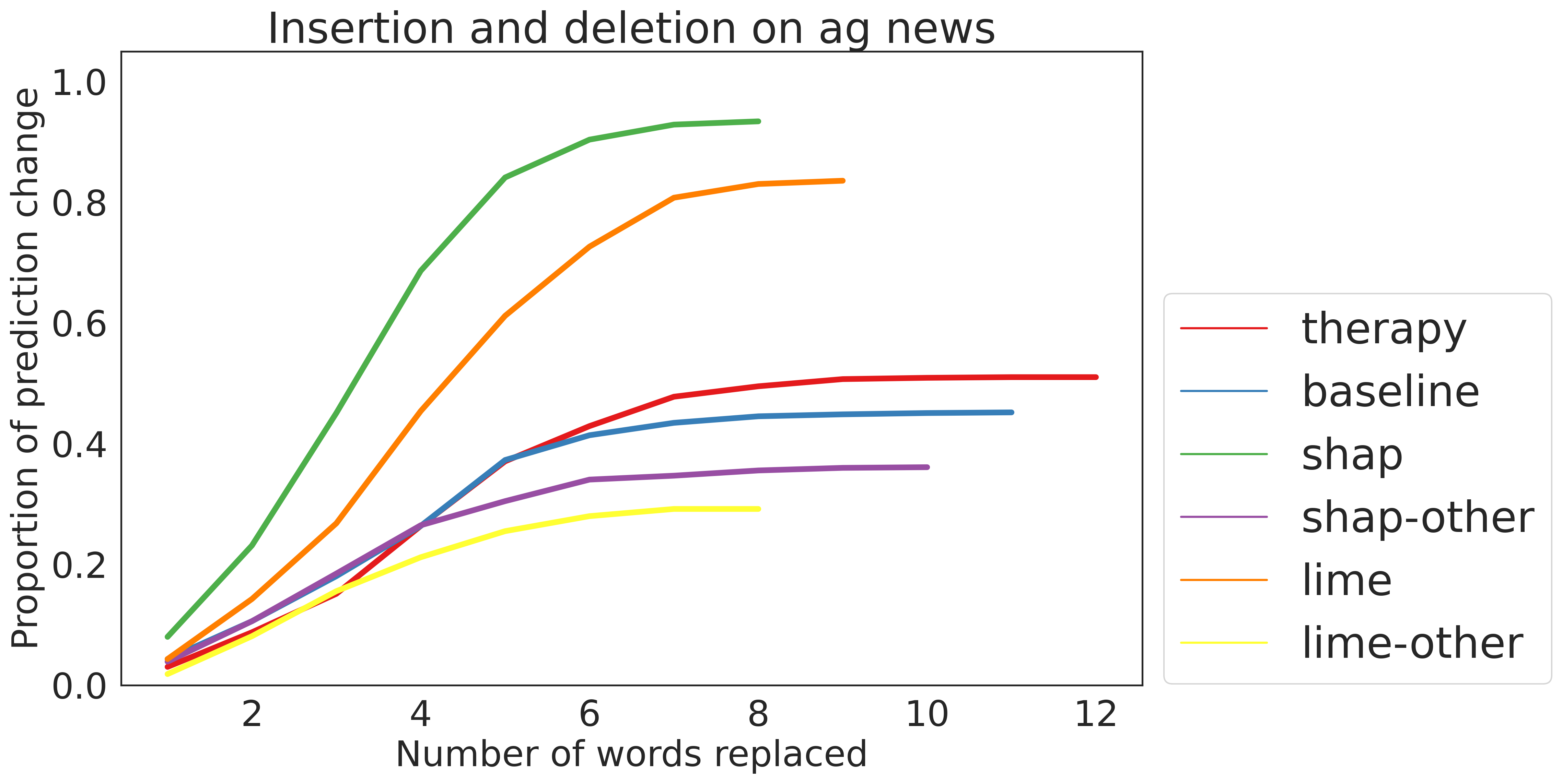}
      \label{fig:test2}
    \end{minipage}
    \caption{Proportion of texts whose glass-box prediction changes w.r.t the number of important words from the original class replaced by important words from other classes.}
    \label{fig:insert_delete}
\end{figure*}

Therapy yields worse results than LIME (although achieving better recall on ag\_news) and SHAP on both datasets. Again, when the input data does not necessarily contain the important features for the model (-other), the results collapse and Therapy outperforms both approaches. This limitation is visible by the plateau in recall scores for these methods: they indeed find the important features \textbf{present in the data}, but are limited to these only, setting the upper limit of features that can be found. In practice, biases contained in the model can be subtle enough not to be present in the available data, in which case LIME and SHAP will not be able to detect it. Therapy, on the other hand, obtains good results while using the same generic LM for both datasets, without using any a priori. The method thus provides a very good overview of the model's behavior when no data, or more broadly, when no data representative of the important features of the model is available. In the latter case, Therapy offers a broader search than the one based on existing texts, offering higher recalls. Again, the baseline is competitive against Therapy on amazon\_polarity but is significantly worse on ag\_news. This illustrates that the cooperative generation allows Therapy to better highlight distinct classes when they are more mixed in the LM.

\subsection{Insertion/deletion of keywords}
\label{subsec:insertion_suppression}
A strategy to validate the correctness of the explanation is to remove the features that the explanation method found important and see how the prediction of the model evolves. The intuition behind deletion is that removing the “cause” will force the model to change its decision~\cite{insertion_deletion}. Similarly, adding a word returned by the explanation as important for another class should lower the confidence of the model. Thus, we compute an insertion/deletion metric that measures the proportion of texts whose glass-box decision changes when a word listed as important for the original class is removed and replaced by an important word from another class.
Figure~\ref{fig:insert_delete} shows the results on both datasets for Therapy, the baseline method, LIME, SHAP, and their version using the other dataset as input (-other) on 1000 texts. Replacements are done by iterating over the list of the top 250 words returned by each method for the original class until the decision of the model changes. Replacement can only occur if the word is present within the text and multiple replacements of the same word in a given text are counted as multiple replacements. This explains why each method has a different maximum number of words replaced. Methods that leverage generative models seem to achieve more replacements. We hypothesize that this is because they are designed to globally explain the model on the input domain, unlike local methods that can return words that are specific to a given input and not generalize well.


We observe that Therapy achieves very similar results to those of LIME and SHAP on amazon\_polarity but significantly worse than both on ag\_news. However, when compared to the -other versions, Therapy achieves very convincing results showing once again that these methods require very specific data while Therapy is able to find important words without accessing any data nor using any a priori on the model. 
In this experiment as well, Therapy outperforms the baseline on both datasets, although the difference is more noticeable on ag\_news.

\section{Conclusion}
Usual explainability methods heavily rely on input data, which is not necessarily available and might not contain model biases or important features. We propose Therapy, a method that leverages cooperative textual generation to create synthetic data that follow the studied model distribution. Thus, the search is driven by a pre-trained LM rather than input samples. The pre-trained LM allows a broader exploration than being restricted to input data neighborhood, relaxing most of the constraints and a priori induced by examples-driven methods. In the extreme case where extremely representative data (such as the test set of a given dataset) of important features of the model is available, Therapy lacks a bit behind state-of-the-art SHAP while being competitive. However, when considering more realistic cases where we do not explicitly give the important features to the explainer or do not have any available data, its performances are very good whereas the other methods are collapsing when even applicable.
Comparisons with a generate-then-classify baseline highlight the benefits of the cooperative generation when the LM does not generate texts that are representative of a single specific class by itself. 
Therefore, Therapy is a useful tool to explore the model behavior on a large domain when collecting data that exactly match the downstream distribution is not feasible. 
Finally, we opposed the proposed approach to LIME and SHAP to highlight the interest of generating representative texts using cooperative generation when input data is lacking. However, an interesting avenue of research would be to use these established explainability methods on cooperatively generated texts, replacing the proposed logistic regression on the tf-idf representations. This potential combination might allow to leverage their performance while alleviating the input data dependency.
\bibliography{anthology,custom}
\bibliographystyle{acl_natbib}

\appendix

\section{Qualitative results}
\label{sec:qualitative_results}
In this appendix, we provide samples generated by Therapy as well as the first 20 top words returned by the method for the different classes of both datasets. Please note that some "words" correspond to sub-words, due to the breakdown into unigrams (ve, ll, ...). The proposed approach allows Therapy to study the impact of n-grams, but this is not possible with LIME and SHAP (using available code), so we restricted the study to unigrams.

\subsection{amazon\_polarity, "positive" class}
Samples:
\begin{itemize}
    \item \texttt{"The amount of support they gave was amazing. I'm so happy with the game and it's a great way for me to play my favorite role!  Also, I love that they're giving us a new character every week, which is awesome :)I think they've been pretty good lately too. It's nice when we see someone who really loves their team and has fun doing things"} 
    \item \texttt{"The whole point of this post is to show people what a great job you've put in.  You are doing an excellent job, I love this sub.  Keep going.  :)I'm so happy for all those people that have helped! It's amazing to be here and it was fun being the one that got the most attention on the other day :). Thank you so much <3You're very welcome :)  And yes... I am glad you enjoyed this subreddit - it is a great place"} 
\end{itemize}

Top-words: \textit{great, love, good, ve, years, people, lot, friends, fun, life, world, works, easy, things, happy, heard, including, awesome, nice, family}

\subsection{amazon\_polarity, "negative" class}

Samples:
\begin{itemize}
    \item \texttt{"I guess you could say it has to be the most boring thing ever.It is a very boring experienceI don’t think they have any idea how to make money off this, so they will probably just do what you say they will if it doesn't work. They may try to sell you on the fact that it isn’t working and then try and get you to buy back into it when they are done making money. If I was them, would they not want to pay me?" }
    \item \texttt{"It might be a bug, but it doesn’t work on my phoneYeah that sucks! I was hoping the same thing happened.I don't think you can get a refund without having to buy a new one...You have no idea what they would be like without the warranty.  They wouldn't know how much money is left if they didn't do it properly or at all and they would probably just give it back as soon as they found out it was defective (or something)"}
\end{itemize}

Top-words: \textit{don, money, bad, doesn, didn, idea, work, device, isn, thing, guess, wrong, back, buy, fact, time, phone, point, problem, thought}

\subsection{ag\_news, "world" class}
Samples:
\begin{itemize}
    \item \texttt{The world's most popular and sought after luxury home in Dubai, the Al Khaleel International Airport (ALI), is a city that is renowned as the hub of the Middle East and North Africa region. It is also the capital city of Abu Dhabi, United Arab Emirates with a population over 2 billion people and a GDP per person.Al-Khaliel airport is an important gateway into Europe to the north east and west of Europe, Asia, and Australia}
    \item \texttt{In the last month, the US government has been accused of using a “secret” military intelligence agency in Iraq and Afghanistan for spying on its allies, including Pakistan, Turkey, Iran, Russia, Syria, Sudan, Lebanon, Libya, Egypt, Nigeria…and even Saudi Arabia!According to a new report published today, the Pentagon is now investigating the alleged use by the US military Intelligence Community (IUC), which was created by President Bush after the 9/11 terror attack that killed over}
\end{itemize}

Top-words: \textit{people, man, country, city, party, killed, family, agree, wrong, general, children, sex, president, police, working, military, dead, missing, woman, days}

\subsection{ag\_news, "sport" class}
Samples: 
\begin{itemize}
    \item \texttt{I think it's pretty safe to assume that the guy has a lot of experience with the game. He played in the NHL, and he was a top 10 player on the team for most games last season (he had a goal in his final three playoff series), and he won a Stanley Cup as a rookie this past season (he finished third in the league in points scored, which was good for second in the league) and is still one of the best players in hockey at this stage in the year}
    \item \texttt{We've got to keep playing this game. This team needs to win games and we need to play the best basketball that's been in our league all season, every night. And it is time to get out there and do that.""The Warriors have won three straight games at home and are 2-0 against the NBA’s best team on both sides of the ball this season with a record of 21-1 (13.7 points per game)}
\end{itemize}

Top-words: \textit{time, game, back, season, play, didn, team, guy, field, night, games, left, 12, title, won, saturday, playing, great, day, wasn}

\subsection{ag\_news, "business" class}
Samples:
\begin{itemize}
    \item \texttt{I am still in shock after hearing of that.It's a pretty big deal. It happened last month. They are trying to get the money out of the company by selling their stock for profit so they can sell more shares and buy more shares at higher prices (which I think would have helped with the stock market) and it was reported as an ""investment fraud"" by the SEC which has been going on all over this subreddit for months, but no one ever seems to care much}
    \item \texttt{Biden is planning to spend millions of dollars to buy a new home, but the real estate market in America is still struggling with the housing shortage. The average house sale cost \$1 billion and was up by nearly 50 percent from the previous year’s price of about \$800 million — according to the Real Estate Board of New York (RBE).The RBE estimates that the average house sales prices are expected to rise 1,000 per month this fiscal year as the economy continues its rebound}
\end{itemize}
Top-words: \textit{money, buy, care, doesn, things, deal, pay, worth, business, car, biggest, interested, month, trade, don, compagny, happened, store, kind, price}

\subsection{ag\_news, "sci/tech" class}
Samples:
\begin{itemize}
    \item \texttt{2K Games' Dark Souls 3 is coming to PC, Mac \& Linux in the near future.The new game will launch for free on PC, Mac \& Linux and Xbox One, PlayStation 5 and Microsoft Windows, as well. It'll come out sometime during this week, with an official release expected soon thereafter, though we don't yet know what it will be called or where exactly you're getting the title. We also have some news from Sony that's not quite so surprisingetc...}
    \item \texttt{In this new age of technology, the world needs more people. We have a lot in our hands. The internet can help us connect to others through video chat and online games.""The company will launch a mobile game called 'Gangster', where it plans to offer ""an interactive experience"" with its users, according to the company. The game has been developed for the Apple iPad and Android phones that use Apple TV, which also uses Google Chromecast, according to a release.}
\end{itemize}
Top-words: \textit{ve, ll, idea, phone, internet, make, system, video, online, life, understand, version, pc, found, 13, thing, computer, lot, hard, issue, people, work, information, future}

\end{document}